\documentclass[letterpaper]{article} 
\usepackage{aaai24}  
\usepackage{times}  
\usepackage{helvet}  
\usepackage{courier}  
\usepackage[hyphens]{url}  
\usepackage{graphicx} 
\usepackage{natbib}  
\usepackage{caption} 
\usepackage{algorithm}
\usepackage{algorithmic}

\usepackage{eucal}
\usepackage{multirow}
\usepackage{arydshln}
\usepackage{enumitem}
\usepackage{xspace}
\usepackage{amsmath}
\usepackage{booktabs}
\usepackage[T1]{fontenc}
\usepackage{xcolor}
\usepackage{caption}
\usepackage{subcaption}

\usepackage{newfloat}
\usepackage{listings}
\DeclareCaptionStyle{ruled}{labelfont=normalfont,labelsep=colon,strut=off} 
\floatstyle{ruled}
\newfloat{listing}{tb}{lst}{}
\floatname{listing}{Listing}
\newcommand{\proj}{TimeGraphs\xspace}

\newcommand{\xhdr}[1]{\vspace{2mm}\noindent{{\bf #1.}}}

\newcommand{\hide}[1]{}
\newcommand{\tami}[1]{}
\newcommand{\tamihide}[1]{}

\title{\proj: Graph-based Temporal Reasoning}
\author{Paridhi Maheshwari$^\dagger$, Hongyu Ren$^\dagger$, Yanan Wang$^\ddagger$, Rok Sosi\v{c}$^\dagger$, Jure Leskovec$^\dagger$}
\affiliations{%
  $^\dagger$Stanford University, $^\ddagger$KDDI Research \\
  {\tt {\{paridhi,hyren,rok,jure\}@cs.stanford.edu, wa-yanan@kddi.com}}
  }

\usepackage{bibentry}

\makeatletter
\renewcommand{\ps@plain}{%
  \renewcommand\@oddfoot{\hfil\thepage\hfil}%
  \renewcommand\@evenfoot{\@oddfoot}%
}
\makeatother

\begin{document}

\maketitle
\begin{abstract}
Many real-world systems exhibit temporal, dynamic behaviors, which are captured as time series of complex agent interactions.
To perform temporal reasoning, current methods primarily encode temporal dynamics through simple sequence-based models.
However, in general these models fail to efficiently capture the full spectrum of rich dynamics in the input, since the dynamics is not uniformly distributed.
In particular, relevant information might be harder to extract and computing power is wasted for processing all individual timesteps, even if they contain no significant changes or no new information.
Here we propose TimeGraphs, a novel approach that characterizes dynamic interactions as a hierarchical temporal graph, diverging from traditional sequential representations. Our approach models the interactions using a compact graph-based representation, enabling adaptive reasoning across diverse time scales. Adopting a self-supervised method, TimeGraphs constructs a multi-level event hierarchy from a temporal input, which is then used to efficiently reason about the unevenly distributed dynamics. This construction process is scalable and incremental to accommodate streaming data.
We evaluate TimeGraphs on multiple datasets with complex, dynamic agent interactions, including a football simulator, the Resistance game, and the MOMA human activity dataset. The results demonstrate both robustness and efficiency of TimeGraphs on a range of temporal reasoning tasks.
Our approach obtains state-of-the-art performance and leads to a performance increase of up to 12.2\% on event prediction and recognition tasks over current approaches. Our experiments further demonstrate a wide array of capabilities including zero-shot generalization, robustness in case of data sparsity, and adaptability to streaming data flow.
\end{abstract}

\section{Introduction}\label{sec:intro}

\begin{figure*}[!h] 
\centering
\includegraphics[width=0.9\textwidth]{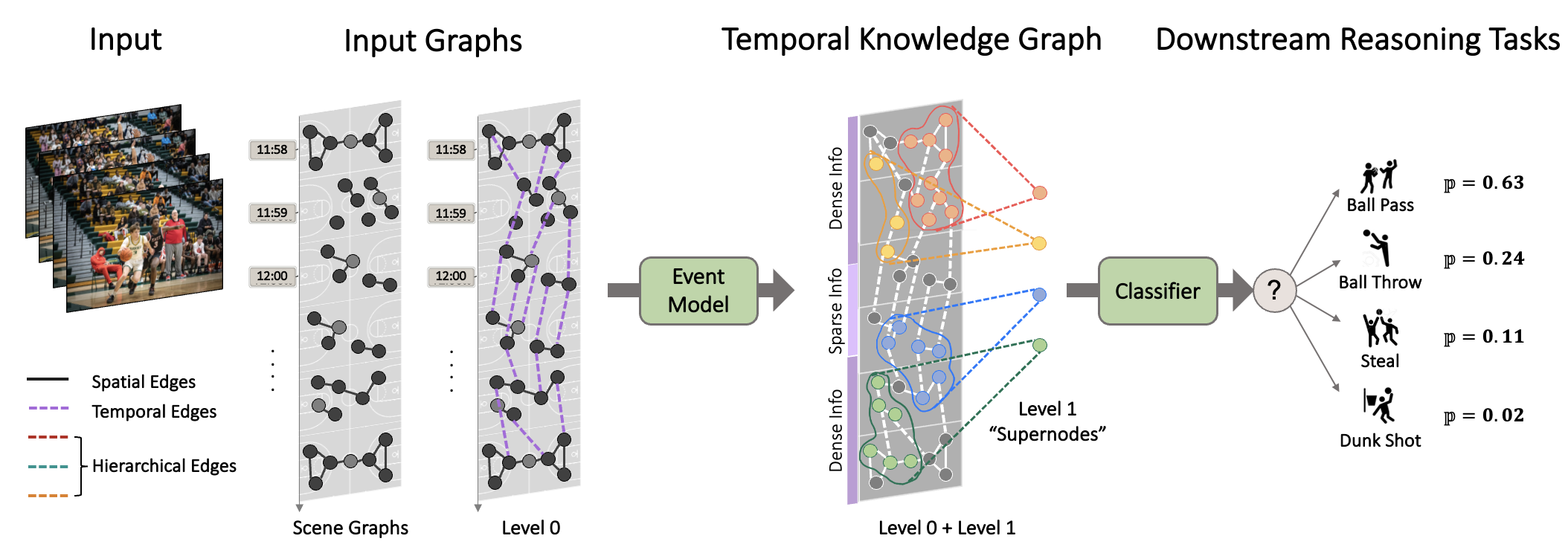}
\caption{\textit{\proj Architecture}: We take as input frame-wise scene graphs (obtained from video or sensors). We add spatial as well as temporal connections to obtain level 0 of our temporal knowledge graph. Next, we use our event model to create level 1 hierarchical events that can uncover latent dependencies at multiple scales. Finally, we leverage the rich temporal knowledge for various downstream reasoning tasks, such as event recognition, using a separate classifier module.
}
\label{fig:timegraphs}
\vspace{-3mm}
\end{figure*}


Temporal sequences of graphs have been increasingly used in many domains to study complex behaviors and interactions.
Such sequences can be found in areas such as 
sport analysis with the graph at each timestep representing the state of players and how they interact cooperatively or competitively \cite{xenopoulos2021graph,anzer2022detection}.
Another example can be seen in
transportation studies, with graphs representing traffic flow or public transit networks over time \cite{li2021spatial,jiang2022graph}.
The rise in the graph sequence representation can be attributed to the multifaceted data representation that they provide, encapsulating complex dependencies in time and across entities \cite{li2015gated,hu2019graph}. 
Current methodologies in handling temporal graph sequences often use simple sequence-based models or graph neural networks or a combination of both to learn a representation over the input sequences of graphs \cite{chen2019gated,wang2022sparsification}. In this way, they are able to model both the dependency between variables within a single timestep and that across timesteps \cite{zhou2020graph}.


However, existing models \cite{duan2022multivariate,liu2023todynet} operate under the assumption that the dynamics in the input
is uniformly distributed across time, thus assigning the same modeling power to every individual timestep. This approach results in an inefficient and often inaccurate distribution of the modeling and computation power as the events within a sequence are rarely distributed evenly over time.
Some timesteps may hold a wealth of information, characterized by dense and rapid dynamics, while others might be less informative with slower changes and fewer events. For example, in the context of basketball games, 
there exist complex dependencies in the way the game evolves and develops \cite{kubatko2007starting}. There may be a prolonged period of inactivity where a player holds the ball for a few seconds with no significant impact on the game. Conversely, there may also be a fast-paced multi-person attack on the basket, involving multiple simultaneous events in a short span of time. 
Thus, assigning equal attention to every timestep risks the possibility of overlooking critical information present in highly dynamic timesteps while wasting significant computational resources elsewhere.

Developing a system that can adeptly reason with intricate temporal events holds great promise. By focusing modeling power on time intervals with dense dynamics, we can optimize the system's efficiency and accuracy. Such an advancement would pave the way for crafting more precise models of intricate temporal dynamics, especially in environments like multi-agent systems, enhancing our understanding of how agents interact over time and capturing the nuances of their relationships and behaviors. As a result, our ability to predict and respond to these dynamics would improve significantly, leading to more informed decision-making in various applications.

To address these issues, we propose \proj, a novel graph-based approach for temporal reasoning over complex agent interactions and their behaviors in real-world systems. The key insight of our approach is that we require an adaptive hierarchical modeling to handle the rich, not uniformly distributed dynamics hidden in the input.
\proj achieves this modeling by representing time series in input as graphs and then constructing a hierarchical graph representation, enabling adaptive reasoning across different time scales.
When information is dense, the model learns to navigate time at the most fine-grained level to capture the volatile dynamics.
For timesteps exhibiting consistent dynamics and lesser information, it ascends to a higher level in the hierarchy, allowing for efficient and adaptive reasoning. This dynamic allocation of computational resources enables \proj to adapt to the unique challenges posed by the temporal input graph sequence. Our hierarchical approach also allows for expressing events at higher abstraction levels. Such a hierarchical model allows us to detect both short-term and long-term events in a scalable way by identifying temporal dependencies at multiple scales.

\proj uses a self-supervised graph pooling method to build the event hierarchy (Figure~\ref{fig:timegraphs}). This graph pooling module, termed \textit{event model}, is trained using a novel contrastive objective based on mutual information maximization \cite{tschannen2019mutual}. By optimizing the mutual information, \proj can uncover the underlying structure in the data, thereby autonomously learning to create and traverse the hierarchy. Once the event model is trained, it can be used to build the hierarchy of events from the input sequences of graphs, resulting in the temporal knowledge graph. The resulting graph is then used to train a graph neural network for downstream applications. An important characteristic of our approach is that it operates in a streaming fashion, reading one input graph at a time.


We illustrate the capabilities of \proj on several datasets for complex temporal reasoning tasks. All dataset involve reasoning about videos and classifying them into a pre-defined set of event categories. We consider predicting outcomes in complex multi-player games that feature dynamical aspects of adversarial groups such as football games from Google's simulator~\cite{kurach2020google}, and Resistance games which comprise of evolving face-to-face interactions in a multi-player social game~\cite{bai2019predicting,kumar2021deception}. Lastly, we apply \proj to the video-based human activity classification dataset Multi-Object Multi-Actor Activity Parsing (MOMA)~\cite{luo2021moma}. Our results show that \proj outperforms sequence-based methods and achieves up to 7\% improvement on these diverse datasets.
With our novel methodology, \proj presents a more efficient way of dealing with graph sequences. By adaptively allocating modeling power and utilizing self-supervised learning, it not only addresses the inherent limitations of existing methods but also introduces a new way of modeling temporal graph sequences.

\section{Method}\label{sec:method}

\subsection{Temporal Reasoning}
Without any loss in its generality, we describe our method in the context of temporal reasoning over a sequence of graphs. This domain is highly demanding, offering a wide range of datasets and benchmarks for the development of new methods that deal with temporal data in multiple modalities, e.g, sequence of structured scene graphs in a video. 
Given a sequence of input graphs, we can obtain the graph $G_t = (\mathcal{V}_t, \mathcal{E}_t)$ at time $t$, where $\mathcal{V}_t$ is the set of nodes and $\mathcal{E}_t$ is the set of edges between the nodes. Nodes and edges can also have associated features.

Given the input, our aim is to develop a knowledge representation that captures important temporal patterns and behaviors and that can be used for temporal reasoning tasks, such as event prediction (binary classification for whether or not an event has occurred) or event recognition (multi-class multi-label classification into different categories of events). Our approach for temporal reasoning works as follows. First, we use self-supervised pre-training to construct a temporal knowledge graph with hierarchical events. This pre-training step is task agnostic. Second, we use the temporal knowledge graph to train a graph neural network for specific downstream tasks (see Figure~\ref{fig:timegraphs}).

\subsection{Temporal Knowledge Graph}
We call our central representation temporal knowledge graph. It consists of many connected subgraphs, where each subgraph represents an event, an important spatial-temporal pattern in input data. The nodes in events have associated time stamps and represent entities, which can be humans or objects, and the edges encode relationships between the entities. The edges can connect nodes within one event or between different events, indicating temporal relationships. Nodes and edges can have additional associated features. These relations between events and their nodes allow our approach to identify relevant past events, including those that might possibly occur in the distant past. Furthermore, temporal knowledge graphs can represent hierarchical events, where events at a higher level are connected to events at a low level. The aim is that events at higher levels of hierarchy capture increasingly more complex patterns and behaviors, spanning a longer time interval and involving more entities, allowing for representation and reasoning at a higher level of abstraction. A graph-based representation provides many benefits: (i) Complex dependencies in time can naturally be modeled via temporal graphical relations; 
(ii) Dynamics are not evenly distributed and hence require a hierarchical approach to adaptively learn the representation of the input from different levels of abstraction;
(iii) The possible branching realizations of the future can then be captured as a distribution over such trees/graphs.
Our hierarchical representation allows us to handle both short-term and long-term events in a scalable way by learning from repeating patterns in data and identifying temporal dependencies at multiple scales. Given a set of input graphs $\{G_{t_1},\cdots ,G_{t_n}\}$, we thus aim to construct a temporal knowledge graph, capturing important events.


First, we use the input graphs to construct the base level of the temporal knowledge graph.
The process of constructing this level is relatively straightforward. Each input graph forms a level 0 event in the temporal knowledge graph, preserving the nodes and edges of the input graph.
We further augment the graph by adding temporal edges that connect the nodes of similar entities in consecutive input graphs. This base level can be constructed in a streaming fashion where the output graph is updated incrementally with each new time-stamped graph $G_{t_i}$.

\subsection{Pre-training Hierarchy-aware Event Model}

A major challenge in constructing higher level events in the temporal knowledge graph is that the ground-truth does not exist for those events and therefore direct supervision is not feasible for this learning task. We address this challenge with a self-supervised approach that exploits the graph structure by using similarities and dissimilarities within the data itself. The key goal of this pre-training phase is to discover and represent the latent structure of the graph at the lower level.

In order to produce multiscale hierarchical events, we leverage a graph-pooling method based on mutual information, called vertex infomax pooling (VIPool)~\cite{li2020graph}. The general idea of graph pooling is to pick a certain proportion of nodes and connect them as in the original graph. Although downscaling is bound to lose some information, the goal is to minimize information loss such that the pooled graph can maximally represent the original graph. The VIPool method uses mutual information estimation and maximization to preserve nodes which have high mutual information with their respective neighborhoods. This technique of selecting nodes that accurately represent their local subgraphs results in the most informative subset of nodes.

Mathematically, the task is to identify a subset of nodes $\Omega$ that maximizes the information content -- as defined by the criterion $C(\cdot)$ to measure the node's ability to express neighborhoods:
\begin{align*}
    \max\limits_{\Omega \subset \mathcal{V}} C(\Omega), \text{ such that } |\Omega|=K
\end{align*}

The criterion $C(\cdot)$ has a contrastive formulation where the affinity between node and its neighborhood is minimized and that between node and an arbitrary neighborhood is maximized. We have
\begin{align*}
C(\Omega) \ = \ & \frac{1}{|\Omega|} \sum_{v \in \Omega} \log \sigma \big( T_w({\bf x}_v, {\bf y}_{\mathcal{N}_v}) \big) \\
& + \ \frac{1}{|\Omega|^2} \sum_{(v,u) \in  \Omega} \log \Big( 1 - \sigma \big( T_w({\bf x}_v, {\bf y}_{\mathcal{N}_u} ) \big)  \Big)
\end{align*}
where $\mathcal{N}_v$ represent the neighborhood of node $v$ -- all nodes whose geodesic distance to $v$ is below a certain threshold, $w$ denotes the trainable parameters and $T_w$ is a neural network that estimates the mutual information between a given node and a set of nodes. Specifically, $T_w(\mathbf{x}_v, \mathbf{y}_{\mathcal{N}_u}) = \mathcal{S}_w  \left( \mathcal{E}_w(\mathbf{x}_v), \mathcal{P}_w(\mathbf{y}_{\mathcal{N}_u} ) \right)$ where networks $\mathcal{E}_w(\cdot)$ and $\mathcal{P}_w(\cdot)$ embed nodes and neighborhoods respectively, and $\mathcal{S}_w(\cdot, \cdot)$ is a similarity measure. These components are implemented as follows -- $\mathcal{E}_w$ and $\mathcal{S}_w$ are MLPs and $\mathcal{P}_w$ aggregates nodes features via connected edges.

$C(\cdot)$ thus captures the intuition that selected nodes must have a maximal coverage of their neighborhoods alone. The optimization problem is unfolded using a greedy approach -- select a vertex with the highest $C(\Omega)$ for $|\Omega|=1$ and repeat it sequentially til the desired number of unique nodes is reached. For computational efficiency, the second term in $C(\Omega)$ is approximated using negative sampling of neighborhoods $\mathcal{N}_u$. Finally, the training loss is $\mathcal{L}_\text{hierarchy} = -C(\Omega)$ for learning parameters $w$ via gradient descent.

We achieve multi-scale feature learning using Graph Cross Networks~(GXN)~\cite{li2020graph}. It comprises a pyramid structure where pooling and unpooling operations are repeated sequentially to get multiple levels in a hierarchy and loss aggregated across levels. Further, feature-crossing between levels enables information flow between coarser and finer representations of the graphs. Finally, deep features from all scales are combined in a multi-scale readout, providing hierarchy-aware reasoning on downstream tasks.

\subsection{Constructing Higher Order Events}

Given the pre-trained event model, we construct higher level events as follows. We pass level 0 of temporal knowledge graph $G$ through the event model to obtain a subset of nodes $\Omega$.
These nodes serve as seeds for the nodes at the higher level of $G$.
For every node in $\Omega$, we then add a new higher-level node, called a \textit{supernode}, to $G$. These supernodes form the next hierarchical level of $G$. The features of supernodes are the aggregate of their local neighborhoods at the lower level. We connect the supernodes with two types of edges: (i) hierarchy edges that connect supernodes at level $l$ to its neighbors at level $l-1$, denoting information relations across levels, and (ii) connections between supernodes themselves denoting information relations at the higher level. This process can be iteratively repeated to construct additional hierarchical levels on top of new supernodes. The outcome is the fusion of graphs from \textit{all} scales/levels of the hierarchy in a single representation, which prevents us from losing information by utilizing only a particular level and allows us to create a richer temporal knowledge graph structure.

\subsection{Fine-tuning for Downstream Tasks}

The temporal knowledge graph constructed in a hierarchical fashion can now be fine-tuned for downstream applications. We consider the task of event prediction and classification which requires temporal reasoning across various time horizons. In particular, the graph-level prediction task takes a sequence of time frames as input and classifies it into a set of event categories.

To achieve this, we train a graph neural network on the temporal knowledge graph which contains higher-order events from the time window. Note that our approach is general and can be coupled with any graph network. In our experiments, we use relational-GCN~\cite{schlichtkrull2018modeling} as our data consists of relational edge types (such as spatial, temporal and hierarchical). Additionally, we incorporate extra node-level features from the trained hierarchy constructor, such as attention scores and one-hot vectors for identifying the level of each node. This is followed by a classifier head to predict event labels. The downstream model can then be finetuned using the binary cross-entropy loss across labels, as the standard choice for the task. Mathematically,
\begin{align*}
\mathcal{L}_\text{classification} = - \frac{1}{|\mathcal{C}|} \sum\limits_{c \in \mathcal{C}} p_c \mathbf{y}_c \log \mathbf{\hat{y}}_c + (1-\mathbf{y}_c) \log (1-\mathbf{\hat{y}}_c)
\end{align*}
where $\mathcal{C}$ denotes the set of event categories and $p_c$ is the weight of a positive sample determined from label distribution for that event. While this is one application that can be advanced using our \proj method, we believe our approach to build a richer temporal knowledge graph is general and other applications can be benefit as well.

\subsection{Multi-task End-to-end Learning}

We have introduced a two-phase approach where we first employ self-supervised pre-training of the event model and then fine-tune on the downstream classification task. Alternatively, we can also set up a multi-task problem where the event model and classifier head are trained together. For this end-to-end (E2E) approach, we adaptively weigh the two losses as 
\begin{align*}
    \mathcal{L}_\text{e2e} = \mathcal{L}_\text{classification} + \alpha \mathcal{L}_\text{hierarchy} \; \text{ where } \; \alpha = 2 - \frac{\text{epoch}}{\text{\# epochs}}
\end{align*}
where hyperparamter $\alpha$ linearly decays with epoch number from $2$ to $1$ during training. This setup gives higher importance to graph-pooling loss in initial iterations and gradually increases the weight of classification loss with epochs, thereby balancing the two tasks.

\section{Experiments and Results}\label{sec:experiment}

\begin{table*}[!h]
\centering
\begin{tabular}{l p{1.3cm} p{3.3cm} p{1.8cm} p{4.4cm} c}
\toprule
\multicolumn{1}{c}{\textbf{Dataset}} & \multicolumn{1}{c}{\textbf{Nodes}} & \multicolumn{1}{c}{\textbf{Node Features}} & \multicolumn{1}{c}{\textbf{Edges}} & \multicolumn{1}{c}{\textbf{Identified Event Categories}} & \multicolumn{1}{c}{\textbf{Window Size}} \\
\midrule
Football & \{players\}, ball & positions, velocity, team identity, player roles, tired factor & spatial; ball & scoring; shooting; ball passing; ball possessioan; goal kick; free kick; corner; throw in & 9 sec (5 frames) \\
Resistance & \{players\} & facial action unit, speaking probability, emotions & interaction prob.; global & outcome (success/failure) & 33 sec (100 frames) \\[0.1ex]
\hdashline\noalign{\vskip 0.6ex}
MOMA & \{actors\}, \hspace{10mm} \{objects\} & groundtruth actor or object class & spatial-temporal & 17 activity classes; 67 sub-activity classes; 52 atomic action classes & \multirow{2}{*}{\parbox{1.7cm}{\centering up to 175 sec (175 frames)}} \\
\bottomrule
\end{tabular}
\caption{Graph formulation for all datasets. Note that we formulate certain features (e.g. team identity and player role which are non-binary) as one-hot vectors. If a feature is not applicable to a certain node type (e.g., tired factor for ball), we set it to zero.}
\label{datasets}
\vspace{-3mm}
\end{table*}

In this section, we evaluate the performance of \proj on the following temporal reasoning tasks: event prediction and action recognition. For event prediction, the task is to predict the next event in the immediate future given a sequence of frames. In action recognition, the task is to classify a sequence into a pre-defined set of categories. The categories could be complete events (e.g., `goal kick') or simpler actions (e.g., `walking'). We start by describing our dataset curation and then proceed to experimental setup and results.

\subsection{Datasets}
We apply our proposed \proj approach to novel datasets featuring complex dynamics of adversarial groups. We describe these datasets along with required pre-processing steps to obtain frame-level graphs:

    \xhdr{Football} We use the Google Research Football simulator~\cite{kurach2020google} to generate full-game football matches. Leveraging the authors' pre-trained models, we create a reinforcement learning environment in which agents are trained to play football in an advanced, physics-based 3D simulation. Each agent or player in the simulation can be assigned different roles: a goalkeeper, defense, offense, or middle field. By following common practices, we choose a different mix of player roles in the simulator and create $6$ teams with varying strategies -- such as focusing heavily on defense or offense and in-between modes. Then we simulate $10$ matches for each pair of teams, resulting in a total of $150$ games. Each game's state, which includes player and ball positions as well as major events, is provided at a rate of $0.55$ frames per second for the entire duration of the game ($90$ minutes). Finally, we create input graphs: nodes represent all the players and the ball, and edges capture their spatial closeness.

    \xhdr{Resistance} We also evaluate performance on a multi-user social game called Resistance (similar to the Mafia game). Each game involves $5-8$ players divided into deceivers and non-deceivers and lasts between $45-60$ minutes over several rounds. 
    By analyzing the outcome (success / failure) of every round, the goal for non-deceivers is to identify the deceivers as early as possible. 
    Our dataset covers $62$ games and $313$ rounds in total, with networks provided at a rate of $3$ frames per second~\cite{bai2019predicting,kumar2021deception}. Each network captures details about individual players (such as facial action unit~\cite{baltrusaitis2018openface}, speaking prediction and emotions) and pairwise features (probability that a player looks at another player). 
    We build the input graphs using the player's attributes as node features and connect the edges by thresholding on the node pair probabilities.

    \xhdr{MOMA} Lastly, we evaluate on a public dataset for video-based human activity classification, namely Multi-Object Multi-Actor Activity Parsing~\cite{luo2021moma}.
    Aiming for hierarchical human activities,
    MOMA provides $373$ raw videos at activity level, annotated in $17$ activity classes (e.g., dinning service). Additionally, each raw video of a larger activity is split into several videos at sub-activity level (e.g., take the order, serve food, clean up the table), which are labeled with one of $67$ sub-activity classes. 
    Given scene graphs of video frames, we build a spatial-temporal graph by introducing temporal context edge to link the same actor/object by tracking them across all frames in the video.

These datasets are a good choice because their constituent events are complex in nature, range over different durations and contain dependencies across entities at varying time scales. For every dataset, we identify relevant event categories for downstream prediction subject to data availability. Additionally, we pick a suitable window size $\Delta t$ for history frames depending on the average duration of events in the dataset or manual observations (see Table~\ref{datasets} for more details). For all datasets, we divide full-length activities into training / validation / test sets using a $7:2:1$ data split.

\subsection{Baselines}
For curated datasets in our work -- namely football and Resistance games, we implement the following baselines and compare their performance with our \proj approach:

    \xhdr{Counts} This is a simple count-based statistical classifier which uses event-category distributions to predict the result based on a random sample from a uniform distribution.
    
    \xhdr{LSTM} Standard deep learning approaches use recurrent architectures for modeling sequence data. Here, we use LSTMs~\cite{hochreiter1997long} that can capture long-term dependencies better. We flatten the features of all nodes in the graph at every timestep, resulting in a fixed-length feature at every time. A sequence of these vectors is input to the recurrent layer, followed by a classification head for predicting the category of event.
    
    \xhdr{Transformer} Given the recent success of modeling attention in a transformer~\cite{vaswani2017attention} model, we also use a transformer encoder on the sequence data. We add a position encoding to the flattened input (from recurrent models) to embed time. In the end, we average the embeddings of all token embeddings (here, every token corresponds to a timestep) and pass through a fully-connected classifier head.
    
    \xhdr{Relational Graph Convolutional Nets} To evaluate the benefits of a purely graph-based approach, we consider RGCN~\cite{schlichtkrull2018modeling} which use only the level 0 of the temporal knowledge graphs as opposed to the hierarchical temporal knowledge graph used by \proj. 

The sequence baselines operate on the input scene graphs from the video, while the RGCN method uses level 0 of the temporal knowledge graph. This setting allows us to address key questions of whether the temporal knowledge graphs are useful and whether hierarchical events are useful.


    \xhdr{HGAP} An action parsing network built on the hypergraph of video data. It consists of 2 sub-networks that unify hypergraph neural networks~\cite{feng2019hypergraph} for modeling frame-level action representation and an action recognition model X3D~\cite{fei2020x3d} for video-level features. Here, the hypergraph is a well-structured frame-level spatial-temporal graph for multi-object and multi-actor scenarios, and nodes extracted from each frame are embedded with image features.
    
    \xhdr{Oracle} A model similar to HGAP with the hypergraph using the ground truth and hense should be more informative about the structure of the input sequence.
  

\subsection{Results}
We evaluate different methods using standard metrics for multi-label classification tasks. Particularly, we use exact match (EM) score (proportion of samples where all labels are identified correctly), macro-averaged accuracy (mAP) (average over per-label accuracies) and lastly, macro averages for F1 score, precision, and recall. Higher numbers are better for all the metrics. Our central evaluation metric is the F1 measure, so we pick the epoch and corresponding model checkpoint which has the highest F1 score on the validation set. Note that all models are trained using Adam optimizer until convergence and we use a learning rate of $0.001$ along with a multi-step scheduler and a batch size of $32$.

\begin{table}[!ht]
\centering
\small
\begin{tabular}{l cccc}
\toprule
\multicolumn{1}{c}{\textbf{Model}} & \textbf{F1} & \textbf{Prec.} & \textbf{Recall} & \textbf{EM} \\ 
\midrule
\multicolumn{5}{c}{\textit{Dataset: Football}} \\
\midrule
Counts & $0.158$ & $0.159$ & $0.158$ & $0.187$ \\ 
LSTM & $0.712$ & $0.649$ & $0.854$ & $0.610$ \\ 
Transformer & $0.706$ & $0.639$ & $\bf 0.901$ & $0.602$ \\ 
RGCN & $0.723$ & $0.664$ & $0.873$ & $0.637$ \\ 
{\bf \proj-TP} & $0.775$ & $0.738$ & $0.841^{ \dagger}$ & $0.670$ \\ 
{\bf \proj-E2E} & $\bf 0.794^{ \dagger}$ & $\bf 0.762^{ \dagger}$ & $0.838$ & $\bf 0.702^{ \dagger}$ \\ 
\midrule
\multicolumn{5}{c}{\textit{Dataset: Resistance Games}} \\
\midrule
Counts & $0.469$ & $0.473$ & $0.465$ & $0.236$ \\ 
LSTM & $0.519$ & $0.521$ & $0.526$ & $0.513$ \\ 
Transformer & $0.482$ & $0.521$ & $0.517$ & $0.494$ \\ 
RGCN & $0.613$ & $0.625$ & $0.601$ & $0.607$ \\ 
{\bf \proj-TP} & $0.642$ & $0.607$ & $\bf 0.712^{ \dagger}$ & $0.519$ \\ 
{\bf \proj-E2E} & $\bf 0.685^{ \dagger}$ & $\bf 0.688^{ \dagger}$ & $0.684$ & $\bf 0.681^{ \dagger}$ \\ 
\bottomrule
\end{tabular}
\caption{Evaluation of different models on event classification task for all datasets. The best results across models are highlighted in boldface. $^{ \dagger}$ denotes the best results within \proj, i.e., default two-phase approach: pre-training and fine-tuning (-TP), or end-to-end (-E2E) training.}
\label{main_results}
\vspace{-3mm}
\end{table}

\xhdr{Event Classification} We can make the following observations from the experimental results on the event classification task (see Table~\ref{main_results}). The count-based baseline performs poorly which indicates task difficulty. The modeling of spatio-temporal video data as a temporal knowledge graph improves the performance as RGCN outperforms other baselines. Finally, \proj, either with a two-phase approach (TimeGraphs-TP) or end-to-end training (TimeGraphs-E2E), achieves the highest F1 score across all datasets, both \proj variants surpassing all other methods and achieving a relative improvement of $10.2\%$ for football and $12.2\%$ for Resistance games on the EM metric. This performance demonstrates the effectiveness of the \proj approach for downstream tasks as well as confirms our premise that hierarchy helps in learning short- and long-term events. An unexpected outcome is that Transformer performs slightly worse than LSTM, which could be because the positional encoding is not sufficient to capture the sequential aspect of data.

\xhdr{End-to-end Training} We can also observe in Table~\ref{main_results} that end-to-end training of the event model and classifier (E2E) improves performance over the two-phase \proj in the case of the football and Resistance datasets. This is likely due to the shared structure between the two tasks, which enables the models to exploit the information from both tasks in a better way. This result affirms our intuition that building an event hierarchy is an important intermediate step. 

\begin{table}[!ht]
\centering
\scalebox{0.9}{%
\begin{tabular}{lcccc}
\toprule
\multicolumn{1}{c}{\multirow{2}{*}{\textbf{Model}}} & \multirow{2}{*}{\parbox{1.2cm}{\centering \textbf{Activity Class.}}} & \multirow{2}{*}{\parbox{2cm}{\centering \textbf{Sub-activity Class.}}} & \multicolumn{2}{c}{\textbf{Atomic action}} \\
\cmidrule{4-5}
&&& \textbf{Class.} & \textbf{Local.} \\
\midrule
GCN & $70.0$ & $37.4$ & $29.4$ & $29.1$ \\
HGAP & $73.9$ & $42.5$ & $29.2$ & $30.3$ \\
Oracle & $88.4$ & $44.1$ & $31.3$ & $32.7$ \\
{\bf \proj} & $\bf 95.3$ & $\bf 69.5$ & $\bf 35.8$ & $\bf 44.1$ \\
\bottomrule
\end{tabular}
}
\caption{Evaluating activity classification, sub-activity classification, and atomic action classification and localization on the MOMA dataset using mAP metric ($\%$).}
\label{moma_result}
\end{table}

\xhdr{MOMA Dataset} Here, we compare \proj with strong baseline models using the mean average precision (mAP) metric for activity and sub-activity classification sub-tasks, and multi-label mAP metric for atomic action classification~\cite{simonyan2014two}. The results for MOMA are shown in Table~\ref{moma_result}. Compared with baseline methods, our proposed \proj using temporal knowledge graph significantly improved the Oracle model on activity, sub-activity, atomic action classification, and localization sub-tasks by $6.9\%$, $25.4\%$, $4.5\%$ and $11.4\%$ respectively. Here, the atomic action sub-task further includes a localization sub-task that classifies all atomic actions involved by the actor. 
The results show that our proposed \proj approach utilizing the temporal knowledge graph is effective for capturing fine-grained actor-centric temporal relationship at the object entity level, and also suggests its efficacy for hierarchy-aware human activity parsing tasks in a real-world setting.

\begin{figure*}[!h] 
\centering
\includegraphics[width=0.75\textwidth]{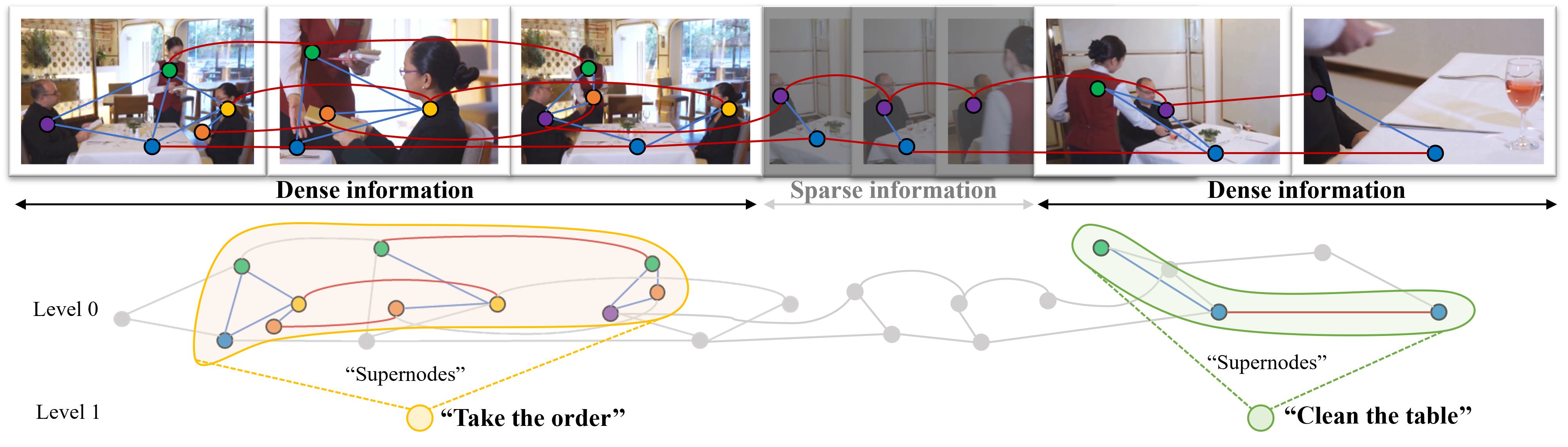}
\caption{TimeGraphs visualization with a MOMA data sample. The proposed event model extracts supernodes at ``level 0: a temporal knowledge graph", which leads to the recognition of ``level 1: hierarchical event".}
\label{fig:vis_timegraphs}
\end{figure*}

\section{Analysis}

In this section, we illustrate additional powerful capabilities of \proj using a range of experiments.

\xhdr{Ablation Study} We ablatively studied the effect of our proposed method on the MOMA dataset in Table~\ref{ablation_result}. By removing the supernodes product by the event model, the performance of sub-activity reduces mAP score by 4.1\%, which suggests the efficacy of our method in modeling hierarchy-aware event representations. Furthermore, we studied how the temporal relationship is crucial for modeling events involving temporal relationships. We compared our method with an ablative model built w/o temporal edges, and the result indicates our method outperforms the ablative model by 2.7\% mAP score. 

\begin{table}[!ht]
\centering
\scalebox{0.8}{%
\begin{tabular}{lccc}
\toprule
\multicolumn{1}{c}{\multirow{2}{*}{\textbf{Model}}}  & \multirow{2}{*}{\textbf{Input graph}} & \multicolumn{2}{c}{\textbf{mAP (\%)}} \\
\cmidrule{3-4}
&& \textbf{Activity} & \textbf{Sub-activity} \\
\midrule
Oracle & hypergraph & $88.4$ & $44.1$ \\
\midrule
{\bf \proj} & spatio-temporal graph & $\bf 95.3$ & $\bf 69.5$ \\
- w/o supernodes & spatio-temporal graph & 95.0 & 65.4 \\
- w/o temporal edges  & spatial graph & 94.8 & 66.8 \\
\bottomrule
\end{tabular}
}
\caption{Ablation study on MOMA dataset. Removing pretrained supernodes reduces sub-activity mAP scores by 4.1\%. Meanwhile, replacing the spatio-temporal graph with a spatial graph reduces 2.7\% mAP.}
\label{ablation_result}
\vspace{-3mm}
\end{table}

\begin{table*}[!ht]
\centering
\small
\begin{tabular}{|c|c|c|c|c|c|c|c|c|c|c|c|c|c|}
\hline
\bf Time (s) & 0 & 50 & 100 & 150 & 200 & 250 & 300 & 350 & 400 & 450 & 500 & 550 & 600 \\
\hline
\bf Acuracy & 33.5 & 52.2 & 59.5 & 53.4 & 53.8 & 57.6 & 61.2 & 62.7 & 65.0 & 65.2 & 64.3 & 65.6 & 65.0 \\
\hline
\end{tabular}
\caption{Performance accuracy as a function of time as a round of the Resistance game progresses.}
\label{tab:mafia_performance_with_time}
\vspace{-4mm}
\end{table*}



\xhdr{Performance over Time} Here, we analyze how prediction accuracy changes with game time. For this task, we consider the Resistance dataset as each data sample represents an independent round within a game. Our input to the model is all frames from the start of a round to the current time. 
That is, we gradually increase the history window, similar to an online model where a stream of data is continuously flowing in. The goal is to predict the outcome of that round. We want to analyze how the prediction accuracy changes as time progresses and the round nears its conclusion. The result is summarized in Table \ref{tab:mafia_performance_with_time} which shows that predictions get better as the game is getting closer to the end. This is justified because there is high uncertainty in the beginning and as the model receives more information, it can make decisions more accurately. Notably, the peaks and drops in the first half of the plot can be attributed to the nature of the game, where certain unseeming moments can affect the outcomes drastically, e.g., when a key player role is revealed. 

\begin{table}[!ht]
\centering
\small
\begin{tabular}{|l|l|l|l|l|l|l|}
\hline
$\Delta t$ (s)          & 5    & 10   & 20   & 40   & 80   & 160  \\ \hline
F1        & 0.76 & 0.65 & 0.52 & 0.40 & 0.34 & 0.31 \\ \hline
Precision & 0.69 & 0.55 & 0.40 & 0.28 & 0.24 & 0.21 \\ \hline
Recall    & 0.85 & 0.80 & 0.74 & 0.68 & 0.59 & 0.58 \\ \hline
\end{tabular}
\caption{Performance when predicting $\Delta t$ seconds into the future on the football dataset.}
\label{tab:football_predicting_future}
\vspace{-3mm}
\end{table}

\xhdr{Predicting the Future} An interesting experiment is to assess how far into the future we can make reasonable predictions using \proj. In particular, our input to the model is a sequence of history frames and the task is to predict an event that is some seconds away, i.e., at frame $\Delta t$ in the future. This is different from the last experiment on performance with time as now our history window is of a fixed length. In our experiments, we vary the future prediction time $\Delta t$ from $0$ frames ($0$ seconds) to $100$ frames ($160$ seconds) to evaluate the model performance as the time horizon gets longer. The trends are summarized in Table \ref{tab:football_predicting_future}. 
The performance drops gradually and monotonically over time, and this is expected behavior as task difficulty increases exponentially with time. 
Notably, this experiment suggests that our \proj approach can be used to make reasonable predictions even for longer time horizons, which is a viable result for developing a real-time prediction system (where lag time is critical) for sports.

\section{Related Work}\label{sec:related}
Several relevant works have been proposed in the field of graph-based temporal reasoning. One such work is MTPool~\cite{duan2022multivariate}, a graph pooling-based framework for multivariate time series classification. TodyNet~\cite{liu2023todynet} is another approach that extracts spatio-temporal features from multivariate time series by learning dynamic graphs and applying temporal convolutions. Additionally, \cite{wang2022sparsification} combines a spatial-temporal graph neural network (ST-GNN) with a matrix filtering module for time series prediction in an end-to-end framework. Similarly to \proj, these works learn graph-based representations of variables and employ hierarchical pooling.
However, these approaches are limited to simple multidimensional time series, while \proj handles graphs of arbitrary shapes and sizes.
Furthermore, \proj uses self-supervision to construct versatile representations of hierarchical events and can be applied to a wide range of tasks, while other approaches are restricted in the sense that they only work for classification tasks.

Another relevant task is analyzing skeleton representations, which comprises the coordinates of key joints within human bodies, with the aim of predicting the corresponding activity. These works~\cite{yan2018spatial,shi2019two} construct spatio-temporal graphs from the frame-level skeleton annotations and employ Graph Neural Networks~(GNNs) for classification. \cite{liu2020disentangling} propose a graph convolution operation (G3D) to facilitate information flow across space and time. MST-GCN~\cite{chen2021multi} develop a subgraph-level network to capture both short and long-term temporal dependencies, but it lacks a hierarchical approach that allows for multi-scale feature learning.
Our work significantly differs from the above approaches as we are not restricted to skeleton data, but use structured graph representations that describe the underlying scene as objects and their relations. 
Additionally, we do not limit to human activity recognition, and classify much more fine-grained and complex scenarios involving multi-player dynamics.

\section{Conclusion}\label{sec:conclusion}
We consider the task of temporal reasoning over input sequences of graphs. We develop and introduce \proj, a novel graph-based architecture that is able to adaptively reason about unevenly distributed dynamics via a hierarchical approach. 
We demonstrate state-of-the-art performance of our approach compared to sequence-based and straightforward graph-based baselines.



\bibliography{main}
\end{document}